%% file: conference_101719.tex
\pdfoutput=1
\documentclass[conference]{IEEEtran}
\IEEEoverridecommandlockouts
\usepackage{cite}
\usepackage{amsmath,amssymb,amsfonts}
\usepackage{graphicx}
\usepackage{textcomp}
\usepackage{comment}
\usepackage{xcolor}
\usepackage{dsbda}

\usepackage{hyperref}
\usepackage{url}

\usepackage{todonotes}

\usepackage{multirow}
\usepackage{booktabs}

\usepackage[ruled,linesnumbered]{algorithm2e}

\input{math_commands.tex}

\renewcommand{\eg}{e.\,g.,\xspace}
\renewcommand{\ie}{i.\,e.,\xspace}
\renewcommand{\etal}{\,et\,al.}

\newcommand{\dataset}[1]{#1}

\newcommand{\numevalsteps}{20,160}
\newcommand{\numexperiments}{1,440}
\newcommand{\embrace}[1]{{(#1)}}

\newcommand{\wos}{\dataset{PharmaBio}\xspace}
\newcommand{\dblpeasy}{\dataset{DBLP-easy}\xspace}
\newcommand{\dblphard}{\dataset{DBLP-hard}\xspace}

\newcommand{\Dt}[1]{\mathrm{tdiff}_{#1}}
\newcommand{\degree}{\mathrm{deg}}
\newcommand{\neighbors}{\mathcal{N}}
\DeclareMathOperator{\tsmin}{\operatorname{time}}

\usepackage{mdframed}
\newmdenv[
  backgroundcolor=gray!20,
  frametitle={Cite as:},
  skipabove=\topsep,
  skipbelow=\topsep,
]{citeas}

\usepackage{amsthm}
\theoremstyle{definition}  
\newtheorem{definition}{Definition}

\newcommand{\node}[0]{vertex\xspace}
\newcommand{\nodes}[0]{vertices\xspace}

\begin{document}

\title{Lifelong Learning of Graph Neural Networks for Open-World Node Classification\thanks{\copyright 2021 IEEE. Personal use of this material is permitted. Permission from IEEE must be obtained for all other uses, in any current or future media, including reprinting/republishing this material for advertising or promotional purposes, creating new collective works, for resale or redistribution to servers or lists, or reuse of any copyrighted component of this work in other works.}}

\author{

\IEEEauthorblockN{Lukas Galke}
\IEEEauthorblockA{
\textit{Kiel University / ZBW}, Germany \\
lga@informatik.uni-kiel.de
}
\and
\IEEEauthorblockN{Benedikt Franke, Tobias Zielke, Ansgar Scherp}
\IEEEauthorblockA{
\textit{Ulm University}, Germany \\
\{benedikt.franke,tobias-1.zielke,ansgar.scherp\}@uni-ulm.de}
}

\maketitle


\maketitle

{\tiny
\begin{citeas}
\begin{verbatim}
@INPROCEEDINGS{galke2021lifelong,
  author={Galke, Lukas and Franke, Benedikt and Zielke, Tobias
          and Scherp, Ansgar},
  booktitle={2021 International Joint Conference on Neural Networks (IJCNN)},
  title={Lifelong Learning of Graph Neural Networks
         for Open-World Node Classification},
  year={2021},
  volume={},
  number={},
  pages={1-8},
  doi={10.1109/IJCNN52387.2021.9533412}
}
\end{verbatim}
\end{citeas}
}

\begin{abstract}
Graph neural networks (GNNs) have emerged as the standard method for numerous tasks on graph-structured data such as node classification. However, real-world graphs are often evolving over time and even new classes may arise. We model these challenges as an instance of lifelong learning, in which a learner faces a sequence of tasks and may take over knowledge acquired in past tasks. Such knowledge may be stored explicitly as historic data or implicitly within model parameters. In this work, we systematically analyze the influence of implicit and explicit knowledge. Therefore, we present an incremental training method for lifelong learning on graphs and introduce a new measure based on $k$-neighborhood time differences to address variances in the historic data. We apply our training method to five representative GNN architectures and evaluate them on three new lifelong node classification datasets. Our results show that no more than 50\% of the GNN's receptive field is necessary to retain at least 95\% accuracy compared to training over the complete history of the graph data. Furthermore, our experiments confirm that implicit knowledge becomes more important when fewer explicit knowledge is available.
\end{abstract}

\section{Introduction}\label{sec:introduction}

Graph neural networks~\cite{DBLP:journals/tnn/ScarselliGTHM09} (GNNs) have emerged as state-of-the-art methods in numerous tasks on graph-structured data such as \node classification\cite{DBLP:journals/corr/KipfW16,DBLP:conf/nips/HamiltonYL17,velickovic2018graph,DBLP:conf/iclr/KlicperaBG19}, graph classification~\cite{DBLP:conf/nips/YingY0RHL18}, link prediction~\cite{DBLP:conf/nips/ZhangC18}, and unsupervised \node representation learning~\cite{velickovic2018deep}.
An intriguing property of GNNs is that they are capable of inductive learning.
An inductive model for graph data only depends on the \node features and the graph structure given by its edges.
In many cases~\cite{DBLP:conf/nips/HamiltonYL17,Xu2020Inductive,DBLP:conf/iclr/ZengZSKP20}, this is a major advantage over models that rely on a static \node embedding~\cite{DBLP:conf/icml/YangCS16}, which would need to be retrained~\cite{perozzi2014deepwalk} as soon as any new \node appears. 
In contrast, inductively trained GNNs can be applied to new data -- or even a different graph -- without any retraining because of the property of only requiring  \node features and edges. 

However, being able to apply the same model to unseen data also comes with challenges that have not been analyzed so far.
Let us assume, we have a \node classification model and new data streams in over time, \ie new edges and \nodes arrive; then even \emph{new classes} may arise!
This raises several new questions:
Do we need to retrain the model? How much past data should be preserved for retraining? Is it helpful to preserve implicit knowledge within the model parameters  or should we retrain from scratch?

\begin{figure}[th!]
    \centering
    \includegraphics[width=\columnwidth]{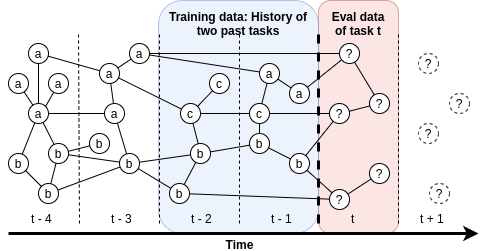}
    \caption{Lifelong Open-World Node Classification. At each time $t$ the learner has to classify new vertices of task $\gT_t$ (red). The learner may use knowledge from previous tasks to adapt to the current task, eventually cut off by a history size (blue). The current task might come with previously unseen classes, \eg see class ``$c$'' that emerged only at task $t-2$ and was subsequently added to the class set. After evaluating each task $\gT_t$, we continue with task $\gT_{t+1}$.}\label{fig:teaser}
\end{figure}

To answer these questions, we frame the problem as an instance of lifelong machine learning~\cite{DBLP:conf/ijcai/ThrunM95,DBLP:conf/kdd/FeiW016,lifelonglearningbook}.
In lifelong learning, the learner has to perform a sequence of tasks $\gT_1,\gT_2, \ldots, \gT_t$, and may use knowledge $\gK$ gained in previous tasks to perform task $\gT_t$ (see illustration in \Figref{fig:teaser}).
In our case, each task consists of classifying \nodes given an attributed graph.
Knowledge $\gK$ may be stored explicitly (the training data of past tasks) or implicitly within the model parameters.
A particular challenge of lifelong learning in the context of graph data is that \nodes cannot be processed independently because models typically take connected \nodes into account. 
We also consider the challenge that the set of classes in task $\gT_t$ differs from classes in previous tasks, which is known as the \emph{open-world classification}~\cite{lifelonglearningbook} problem. 

We address these challenges by introducing a new incremental training method for lifelong learning on graph data.
This training method can be applied on any dataset by the introduction of a new measure to harmonize temporal variances in the evolution of graph datasets with different characteristics and granularities.
The method is capable to integrate various existing GNNs, both isotropic and anisotropic as well as scalable GNN methods.
In our experiments, we thoroughly evaluate representative and scalable GNN architectures (and one graph-agnostic multi-layer perceptron) using our incremental training method that retrains the model for each task.
We use an artificial \emph{history size} that limits the amount of past data (called here: \textit{explicit knowledge}) available for training and compare \emph{limited history-size} retraining against unlimited \emph{full-history} retraining.

Furthermore, we compare reusing model parameters from previous tasks (\emph{warm restart}) against retraining from scratch (\emph{cold restarts}) to analyze the influence of \textit{implicit knowledge}.
In an ablation study, we compare incrementally training models against once-trained models.
To facilitate our analyses, we contribute three new datasets for lifelong learning; one co-authorship and two citation graph datasets with different degrees of changes in the class set.
In total, we have experimented with 48 different incremental training configurations, namely 6 architectures $\times$ 4 history sizes $\times$ cold restarts and warm restarts, which we evaluate on three new datasets. To enable a fair comparison, we tune the hyperparameters for each configuration separately.
We repeat all experiments 10 times with different random seeds.

The results of our \numexperiments{} experiments ($48$ configurations repeated $10$ times on 3 datasets) reveal the following insights:

(i)~\textit{Data from only few past tasks are sufficient for retraining}
Surprisingly, only the data from very few past tasks are sufficient to maintain a high level of accuracy that matches the accuracy of the same model retrained using all past data.  
In our experiments, merely 50\% of the GNN's receptive field (corresponding to history sizes of $3$ or $4$ past tasks) are sufficient to match at least 95\% of the accuracy of the same model trained on the full history.

(ii)~\textit{Reusing parameters requires less data}
Tuning the learning rate may compensate for the knowledge lost when randomly re-initializing the model. However, when the history size is small, using warm restarts tends to be beneficial such that more knowledge from past tasks can be preserved implicitly.
Our results further suggest that using warm restarts is generally preferable for lifelong learning.

(iii)~\textit{Incremental training is a simple yet effective technique to tackle lifelong learning on graphs}
By comparing incrementally trained models with once-trained models, we find that 
the accuracy of once-trained models decreases quickly when new classes appear. 
We note that even when no new classes appear over time, incremental training still benefits from the increased amount of training data.

These insights have direct consequences for using GNNs in practical applications.
It allows to decide how much historical data should be to kept to maintain a certain performance versus having memory available in the GPUs. 
This is an important criterion that influences which GNN methods are applicable~\cite{DBLP:SimpleGCN,DBLP:conf/iclr/ZengZSKP20}.
We publicly provide the datasets as well as the evaluation framework to extend our experiments and accelerate research in lifelong learning on graphs.

Below, we discuss the related works and provide a problem statement with our training procedure and new measure in \Secref{sec:methods}.
We describe the experimental apparatus and datasets in \Secref{sec:apparatus}.
The results of our experiments are reported in \Secref{sec:results} as well as in \Secref{sec:ablation} for our ablation study. We discuss the results in \Secref{sec:discussion} before we conclude.

\section{Related Work}

We discuss the works in lifelong machine learning and especially lifelong learning on graphs. 
Subsequently, we discuss methods for evolving graphs as well as methods regarding history sizes.
GNN architectures will be discussed in more detail in \Secref{sub:models} along with our selection of representative architectures.

\emph{Lifelong learning} (sometimes called: continual learning~\cite{DBLP:conf/nips/Lopez-PazR17}) is present in machine learning research since the mid 1990s~\cite{DBLP:conf/ijcai/ThrunM95,thrun1998lifelong,DBLP:conf/aaaiss/SilverYL13,liu2017lifelong}.
The goal of lifelong learning is to develop approaches that can adapt to new tasks.
Thus, it differs from online learning~\cite{DBLP:conf/icml/HerbsterPW05}, which focuses on efficiency.
ELLA~\cite{DBLP:conf/icml/RuvoloE13} introduces an efficient lifelong learning algorithm with convergence guarantees and employs multi-task learning such that future tasks can improve previous tasks.
Fei et al.~\cite{DBLP:conf/kdd/FeiW016} analyze SVMs in a lifelong learning setting and introduce cumulative learning.
Cumulative learning relates to our approach, as we consider that some data is shared among the tasks. However, we further investigate \emph{how much} past data is necessary to retain accuracy compared to a fully-cumulative approach.
Lopez-Paz \& Ranzato~\cite{DBLP:conf/nips/Lopez-PazR17} introduce a gradient episodic memory framework for the image domain, where examples can be processed independently, and tackle the catastrophic forgetting problem,
\ie the loss of previously learned information when new information is learned~\cite{DBLP:journals/connection/Robins95}.
A recent overview of lifelong learning can be found in~\cite{lifelonglearningbook}.

In the field of \emph{open-world classification}, several methods have been proposed to dynamically detect novel classes for classic machine learning methods~\cite{DBLP:journals/tkde/MasudGKHT11,DBLP:conf/kdd/FeiW016} as well as graph neural networks~\cite{DBLP:conf/icdm/WuPZ20}. For now, we assume that the model is retrained for each task. We lay the foundation for self-detection of new classes by providing our datasets along with a simple yet effective training scheme for lifelong learning on graphs.

Related works on \emph{lifelong learning on graphs} are very limited.
Concurrent with our research, there is one recent work by Wang et al.~\cite{DBLP:journals/corr/abs-2009-00647} that also explores GNNs on lifelong learning problems. 
The authors focus on catastrophic forgetting.
To tackle this challenge, the authors explore a method to cast \node classification as a graph classification task by transforming each \node of the graph into a feature graph. 
This way, the \nodes become independent such that they can follow the streaming setup from \cite{DBLP:conf/nips/Lopez-PazR17}. Apart from lifelong learning, also other methods have been developed to make \nodes independent via preprocessing: SIGN~\cite{signSampling} and Simplified GCN~\cite{DBLP:SimpleGCN}. We include the latter in our experiments.

In the past, also different methods for \emph{evolving graphs} have been proposed including dynamic embedding methods~\cite{DBLP:conf/www/NguyenLRAKK18,lee2020dynamic}, autoencoder-based methods~\cite{DBLP:journals/corr/abs-1805-11273,DBLP:journals/kbs/GoyalCC20}, GNNs for graphs with fixed \node set~\cite{DBLP:conf/icml/TrivediDWS17,DBLP:conf/iconip/SeoDVB18,kumar2018learning,DBLP:conf/iclr/TrivediFBZ19,DBLP:journals/pr/ManessiRM20,DBLP:conf/wsdm/SankarWGZY20,rossi2020temporal}, and inductive GNN methods that can deal with previously unseen \nodes~\cite{DBLP:conf/aaai/ParejaDCMSKKSL20,Xu2020Inductive}.
However, these methods are not relevant to our work 
because we use the time information merely to induce the sequence of lifelong graph learning tasks (see Section~\ref{sub:datasets}). 
When a \node is preserved from task $t$ to $t+1$, it will have also the same features and label(s). 
Thus, our sequence of tasks does not require sophisticated long-horizon temporal information. 
Instead, in our work we require \textit{the ability of the models to adapt to new classes} during a sequence of tasks.

Regarding finding the optimal \emph{history size in data streams}, Fish and Caceres~\cite{fish2017task} treated the window size as a hyperparameter and proposed an optimization algorithm which requires multiple runs of the model. 
This is a rather costly procedure and the study does not yield insights on how much predictive power can be preserved when selecting a near-optimal but much smaller, and thus more efficient, window size. 
Other works, \eg \cite{Ersan2020}, indicate that smaller history sizes might be beneficial in some scenarios.
However, there is no systematic study of the influence of history sizes in lifelong learning on graphs.

To summarize, lifelong learning on graphs is a surprisingly unexplored topic. 
In particular, none of discussed works analyzes the problem of \emph{open-world classification} in graph data and how much past training data is necessary -- or how few is enough -- to maintain good predictive power.

\section{Problem Statement and Methods}
\label{sec:methods}

We first outline our problem formalization of lifelong learning on graphs.
Subsequently, we introduce our incremental training procedure as well as our method to harmonize window sizes across different datasets. Finally, we briefly describe the base models that we incrementally train for our experiments.

\subsection{Problem Formalization}
We define our problem of open-world classification of graph \nodes as a form of lifelong learning~\cite{DBLP:conf/kdd/FeiW016}. It is defined as: 

\begin{definition}[Lifelong Learning~\cite{DBLP:conf/kdd/FeiW016}]
At any time $t$, the learner has performed a sequence of $t$ learning tasks, $\gT_1, \gT_2, \ldots, \gT_{t}$ and has accumulated the knowledge $\gK$ learned in these past tasks. 
At time $t+1$, it is faced with a new learning task $\gT_{t+1}$. 
The learner is able to make use of past knowledge to help perform the new learning task $\gT_{t+1}$.
\end{definition}

We cast this definition into a lifelong graph learning problem by considering each task $\gT_t := (\gG_t, \mX^\embrace{t}, \vy^\embrace{t})$ to be a \node classification task with graph $\gG_t = (V_t, E_t)$, corresponding \node features $\mX^{(t)} \in \mathbb{R}^{|V_t| \times D}$, and \node labels $\vy^{(t)} \in \mathbb{N}^{|V_t|}$.
We denote the set of all classes at time $t$ as $\sY_t$.
To ensure that past knowledge is helpful to perform $\gT_t$, we impose $\gG_{t-1} \cap \gG_t \neq \emptyset$. We assume that the features and labels of the \nodes do not change: $\mX^{(t-1)}_u = \mX^{(t)}_u, \vy^{(t-1)}_u = \vy^{(t)}_u$ if $u \in V_{t-1} \cap V_{t}$. Such changes can still be modeled by inserting a new \node and removing the old one. The task is to predict the class labels for new \nodes $V_t \setminus V_{t-1}$. Please note that these \nodes may come with new, unseen classes as $\sY_t$ may differ from $\sY_{t-1}$.
Furthermore, we analyze the effect of a history size $c$, which limits the available past data. We call this past data \textit{explicit knowledge}. In this case, we set $\tilde{\gT}_t := (\tilde{\gG}_t, \tilde{\mX}^{(t)}, \tilde{\vy}^{(t)})$ with $\tilde{\gG}_t := \gG_t \setminus ( \gG_1 \cup \gG_2 \cdots \cup \gG_{t-c-1})$, and remove corresponding features and labels to construct $\tilde{\mX}_t$ and $\tilde{\vy}_t$. Still, \textit{implicit knowledge} acquired in past tasks, \eg within the model parameters, may be used for task $\tilde{\gT}_t$.

\subsection{Incremental Training for Lifelong Learning on Graphs}\label{sub:incrementaltraining}

Without loss of generality, we assume to have a finite sequence of $T$ tasks $\gT_1,\ldots, \gT_T$ and a model $f$ with parameters $\theta$. 
Over the course of the tasks, the graph changes, including its \nodes, edges, as well as the set of classes.
To address these changes, we explore a simple yet effective incremental training technique for adapting neural networks to new graph-structured tasks.
As preparation for task $\gT_{t+1}$, we retrain $f$ on the labels of $\gT_{t}$ to obtain $\theta^\embrace{t}$.
%
Whenever $l$ new classes appear in the training data, we add a corresponding amount of parameters to the output layer of $f^\embrace{t}$. Therefore, we have $|\theta^\embrace{t}_\text{output weights}| = |\theta^\embrace{t-1}_\text{output weights}| + l$ and $|\theta^\embrace{t}_\text{output bias}| = |\theta^\embrace{t-1}_\text{output bias}| + l$.
Those parameters that are specific to new classes are newly initialized.
For the other parameters, we consider two options in our incremental training procedure: \emph{warm restarts} and \emph{cold restarts}.
With \emph{cold restarts}, we re-initialize $\theta^\embrace{t}$ and retrain from scratch.
In contrast, when using \emph{warm restarts}, we initialize the parameters for training on task $\gT_t$ with the final parameters of the previous task $\theta^\embrace{t-1}$.
Algorithm~\ref{alg:proc} outlines our incremental training procedure.

\begin{algorithm}[h!]
\caption{Incremental training for lifelong graph learning under cold-start vs. warm-start condition \label{alg:proc}}
    \SetKwInOut{Input}{Input}
    \SetKwInOut{Output}{Output}
  \Input{Sequence of tasks $\tilde{\gT}_0, \cdots, \tilde{\gT}_T$, model $f$ with parameters $\theta$, flag for cold or warm restarts}
  \Output{Predicted labels for new \nodes of each task}
  known\_classes $\leftarrow \emptyset$\;
  $\theta \leftarrow$ initialize\_parameters()\;
  \For{$t \leftarrow 1$ \KwTo $T$}{
  new\_classes $\leftarrow \operatorname{set}(\tilde{\vy}^\embrace{t-1}) \setminus \text{known\_classes}$\;
  \If{new\_classes $\neq \emptyset$}{
    $\theta^\prime \leftarrow$ expand\_output\_layer($\theta$, $|\text{new\_classes}|$)\;
  }{}
  $\theta^\prime \leftarrow$ initialize\_parameters()\;
  \If{$t > 1$ \textnormal{\textbf{and}} do\_warm\_restart = TRUE}{
    $\theta^\prime \leftarrow$ copy\_existing\_parameters($\theta$)\;
  }
  
  $\theta^\prime \leftarrow$ train($\theta^\prime$, $\tilde{\gG}_{t-1}$, $\tilde{\mX}^{(t-1)}$, $\tilde{\vy}^{(t-1)}$)\;
  $\tilde{\vy}_\text{pred} \leftarrow$ predict($\theta^\prime$, $\tilde{\gG_t}$, $\tilde{\mX}^{(t)}$) for \nodes $V_{t} \setminus V_{t-1}$\;
  known\_classes $\leftarrow \text{known\_classes} \cup \text{new\_classes}$\; 
  $\theta \leftarrow \theta^\prime$\;
}
\end{algorithm}

\subsection{Compute the k-Neighborhood Time Difference Distribution}\label{sub:deltat}

Real-world graphs grow and change at a different pace~\cite{Aggarwal:2014:ENA:2620784.2601412}. 
Some graphs change quickly within a few time steps like social networks while others evolve rather slowly such as citation networks.
Furthermore, graphs show different change behaviour, \ie different patterns in how \nodes and edges are added and removed over time.
Therefore, depending on the specific graph data, a different history of the data must be used for training to take these factors into account.

We propose a new measure of $k$-Neighborhood Time Difference Distribution $\Dt{k}$, which enumerates the distribution of time differences within the $k$-hop neighborhood of each \node (corresponding to the \emph{receptive field}~\cite{stochasticSampling} of a
GNN with $k$ graph convolutional layers).
This measure ensures that the history sizes are comparable with respect to these factors.

\begin{definition}[$k$-Neighborhood Time Difference Distribution]
Given graph $\gG$ and let $\mathcal{N}^k(u)$ be the $k$-hop neighborhood of $u$, \ie the set of \nodes that are reachable from $u$ by traversing at most $k$ edges. Let $\operatorname{time} : \gV \to \mathbb{N}$ be a function that gives the time information for each \node, \eg the year of a publication.
We define $\Dt{k}(\gG)$ to be the \emph{multiset of time differences}, computed over all vertices $u \in V$ to their $k$-distant neighboring vertices $v \in \neighbors^k(u)$ that occurred before $u$.
\begin{align*}
\label{eqn:deltat}
\Dt{k}(\gG) := &\{ \tsmin(u) - \tsmin(v) \mid \\
            &u \in V \wedge v \in \neighbors^k(u)
\wedge \tsmin(v) \leq \tsmin(u) \}
\end{align*}
\end{definition}

We interpret the multiset $\Dt{k}$ as a distribution over time differences, that is used to further analyze a dataset's temporal distribution (percentiles) and to make datasets comparable.
In our experiments, we compare models trained with a \emph{limited history size} against models trained with the \emph{full history}.
We use the 25th, 50th, and 75th percentiles of this distribution as history sizes versus the 100th percentile to model the full graph to analyze the influence of explicit knowledge. 

\subsection{Models}\label{sub:models}

We select representative GNN architectures as well as scalable GNN techniques for our experiments.
The goal is to understand how different approaches of GNNs react to situation of changing graphs and new classes.
Dwivedi\etal{}~\cite{dwivedi2020benchmarking} distinguish between isotropic and anisotropic GNN architectures.
In isotropic GNNs, all edges are treated equally. Apart from graph convolutional
networks~\cite{DBLP:journals/corr/KipfW16}, examples of isotropic GNNs are
GraphSAGE-mean~\cite{DBLP:conf/nips/HamiltonYL17}, DiffPool~\cite{DBLP:conf/nips/YingY0RHL18}, and GIN~\cite{DBLP:conf/iclr/XuHLJ19}.
In anisotropic GNNs, the weights for edges are computed dynamically. Instances of anisotropic GNNs include graph attention networks~\cite{velickovic2018graph}, GatedGCN~\cite{DBLP:journals/corr/abs-1711-07553}, and MoNet~\cite{DBLP:conf/cvpr/MontiBMRSB17}.
There are further approaches, which have been specifically proposed to scale GNNs to large graphs.
These approaches fall into two categories: 
sampling\cite{DBLP:conf/nips/HamiltonYL17,adaptiveSampling,10.1145/3292500.3330925,DBLP:conf/iclr/ZengZSKP20}, and separating neighborhood
aggregation from the neural network ~\cite{DBLP:SimpleGCN,signSampling,10.1145/3394486.3403296}.
From each of these four categories (anisotropic versus isotropic GNNs and preprocessing versus sampling), we select one representative for our experiments.

We select \textbf{Graph Attention Networks} (GATs)~\cite{velickovic2018graph} as representative for the class of anisotropic GNNs.
In GATs, the representations in layer $l+1$ for \node $i$ are computed as follows:
$\hat{\vh}_i^{l+1} = \alpha_{ii}^l \vh_{i}^l + \sum_{j \in \neighbors(i)} \alpha_{ij}^l \vh_{j}^l$ and $\vh_i^{l+1} = \sigma(\mU^l \hat{\vh}_i^{l+1})$, 
where $\neighbors(i)$ is the set of adjacent \nodes to \node $i$, $U^l$ are learnable parameters, and $\sigma$ is a nonlinearity. 
The edge weights $\alpha_{ij}$ are computed by a self-attention mechanism based on $h_i$ and $h_j$, \ie the softmax of $a(\mU^l \vh_i || \mU^l \vh_j)$ over edges, where $a$ is an MLP and $\cdot||\cdot$ is the concatenation operation.

We select \textbf{GraphSAGE-Mean}~\cite{DBLP:conf/nips/HamiltonYL17} as a representative for isotropic GNNs because its special treatment of the \nodes' self-information has shown to be beneficial~\cite{dwivedi2020benchmarking}.\label{sub:graphsage-mean}
The representations of self-connections are concatenated with averaged neighbors' representations before multiplying the parameters. 
In GraphSAGE-Mean, the procedure to obtain representations in layer $l+1$ for \node $i$ is given by the equations:
$\hat{\vh}_i^{l+1} = \vh_i^l || \frac{1}{\degree_i} \sum_{j \in \neighbors(i)} \vh_j^l$ and $ \vh_i^{l+1} = \sigma(\mU^l \hat{\vh}_i^{l+1})$,

We select \textbf{Simplified GCN}~\cite{DBLP:SimpleGCN}
as a representative for shifting the neighborhood aggregation into preprocessing.  
Simplified GCN is a scalable variant of Graph Convolutional Networks (GCN)~\cite{DBLP:journals/corr/KipfW16} that admits regular mini-batch sampling. 
Simplified GCN removes nonlinearities and collapses consecutive weight matrices into a single one.
Thus, simplified GCN can be described by the equation $\mathbf{\hat{Y}}_\mathrm{SGC} = \text{softmax}(\mathbf{S}^K\mathbf{X}\mathbf{\Theta})$, 
where  $\mathbf{S}$ is the normalized adjacency matrix and $\mathbf{\Theta}$ is the weight matrix. The hyperparameter $K$ has a similar effect as the number of layers in regular GCNs. Instead of using multiple layers, the $k$-hop neighbourhood is computed by $\mathbf{S}^K$, such that $\mathbf{\mS}^K \mX$ can be precomputed.
This makes Simplified GCN efficient, while not necessarily harming the performance~\cite{DBLP:SimpleGCN}.

We use \textbf{GraphSAINT}~\cite{DBLP:conf/iclr/ZengZSKP20} as state-of-the-art subgraph sampling technique.
In GraphSAINT, entire subgraphs are sampled for training GNNs.
Subgraph sampling introduces a bias which is counteracted by normalization coefficients for the loss function.
The authors propose different sampling methods: \node sampling, edge sampling, and random-walk sampling. We use the best-performing random-walk sampling for our experiments.
The underlying GNN is exchangeable, yet the authors suggest to use \textbf{Jumping Knowledge networks} (JKNets)~\cite{DBLP:JKNetwork}. 
JKNets introduce skip-connection to GNNs:
each hidden layer has a direct connection to the output layer, in which the representations are aggregated, e.\,g., by concatenation.
To isolate the effect of GraphSAINT sampling, we also include JKNets in our experiments.

\section{Experimental Apparatus}\label{sec:apparatus}

\subsection{Datasets}\label{sub:datasets}
Pre-compiled graph datasets for lifelong learning are rare~\cite{hu2020open}. 
Many commonly used graph datasets are stripped off from any temporal data.
We contribute three new graph datasets for lifelong learning on the basis of scientific publications: one new co-authorship graph dataset
(\wos) as well as two newly compiled citation graph datasets based on DBLP (\dblpeasy and \dblphard).
For \wos, the classes are journal categories.
For DBLP, we use the conferences and journals of the published papers as classes.
When new conferences and journals emerge, as they do in computer science, new classes will be introduced to the data.
The datasets were generated by imposing a minimum threshold of publications per class per year: 100 for \dblpeasy and 45 for \dblphard, 20 for \wos.
For the co-authorship graph \wos we additionally require a minimum of two publications per author per year.
In all datasets, \node features are normalized TF-IDF representations of the publication title.

\begin{table}[!th]
    \centering
    \caption{Global dataset characteristics: total number of \nodes $|V|$, edges $|E|$, features $D$, classes $|\sY|$ along with \# of newly
    appearing classes (in braces) within the $T$ evaluation tasks}\label{tab:datasets}
    \begin{tabular}{lrrrrrr}
    \toprule
    Dataset & $|V|$ & $|E|$ & $D$ & $|\sY|$ & $T$\\
    \midrule
    \dblpeasy & 45,407  & 112,131 & 2,278 & 12 (4 new)    & 12\\
    \dblphard & 198,675 & 643,734 & 4,043 & 73 (23 new)  & 12\\
    \wos      & 68,068  & 2,1M    & 4,829 & 7  & 18\\
    \bottomrule
    \end{tabular}
\end{table}

\begin{figure}[ht]
    \centering
    \includegraphics[width=\columnwidth]{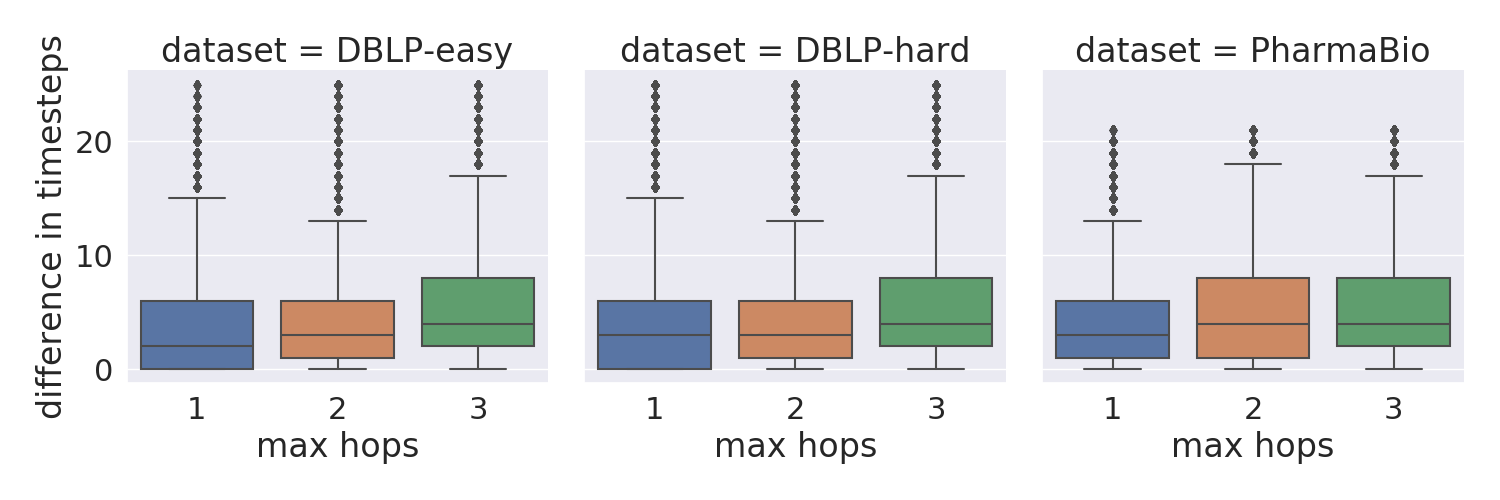}
    \caption{Distributions of time differences $\Dt{k}$ (y-axis) for \dblpeasy (left), \dblphard (center) and \wos (right) within the $k$-hop neighborhood for $k = \{ 1,2,3\}$ (x-axis).}\label{fig:deltat}
\end{figure}

Table~\ref{tab:datasets} summarizes the basic characteristics of the datasets and \Figref{fig:deltat} shows the distribution of time differences $\Dt{k}$ for different values of $k$. We select $1$, $3$, $6$, $25$ as history sizes for DBLP-\{easy,hard\} and $1$, $4$, $8$, $21$ as history sizes for \wos according to the $25$,$50$,$75$,$100$-percentiles of $\Dt{2}$.
For each dataset, we construct the sequence of tasks $\tilde{\gT}_1, \ldots,\tilde{\gT}_T$ on the basis of the publication year along with a history size $c$.
For each task $\tilde{\gT}_t$, we construct a graph with respect to publications from time  $\lbrack t - c, t \rbrack$, where publications from time $t$ are the test \nodes, and $t < c$ training \nodes (transductive). In the inductive case that is used by GraphSAINT in our experiments, we train exclusively on $\tilde{\gT}_{t-1}$, but still evaluate the test \nodes of $\tilde{\gT}_{t}$. We set the first evaluation task $\tilde{\gT}_1$ to the time, in which 25\% of the total number of publications are available.
Thus, mapping the datasets to our problem statement (see Figure~\ref{fig:teaser}), our first evaluation task $t=1$ corresponds to year 1999 in \wos (total range: 1985--2016) and 2004 in DBLP-\{easy,hard\} (1990-2015). We continue with the next years for subsequent tasks.

\begin{figure}
    \centering
    \includegraphics[width=\columnwidth]{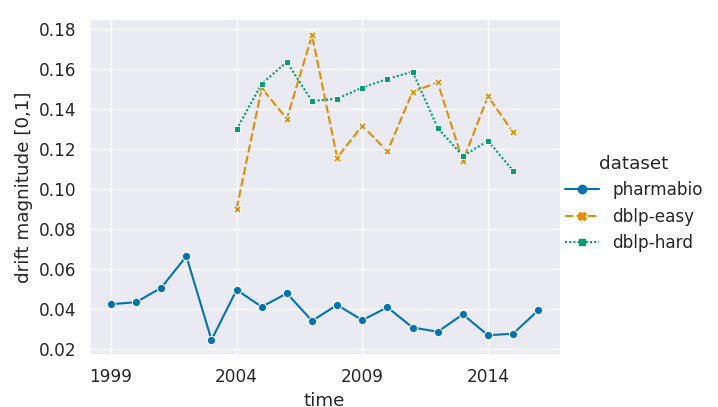}
  \caption{Magnitude of the class drift per dataset. The drift within the \wos dataset (no new classes) is lower than the drift of both DBLP variants. Independent and identically distributed data would have drift magnitude zero.}
    \label{fig:drift_mag}
\end{figure}

Regarding changes in the class set, \dblpeasy has 12
venues in total, including one bi-annual conference and four new venues
appearing in 2005, 2006, 2007, and 2012. \dblphard has 73 venues,
including one discontinued, nine bi-annual, six irregular venues, and 23 new venues.
To quantify changes in the class set, we compute the magnitude of the class drift as total variation distance~\cite{webb2016,webb2018}:

\begin{equation*}
    \sigma_{t-1,t} = \frac12 \sum_{y \in \sY_{t-1} \cup \sY_t} | P_{t-1}(y) - P_{t}(y) |
\end{equation*}
where $P_t(y)$ is the observed class probability at time $t$. 
The drift magnitudes are visualized per dataset in \Figref{fig:drift_mag}.

\subsection{Procedure}

First, we \textit{optimize the hyperparameters} by separately tuning the learning rate for each model, history size, and restart configuration on \dblpeasy
(the effect of weight decay was negligible).
All models are constrained to two graph convolutional layers, a comparable penultimate hidden dimension (2x32 GraphSAGE, 4x8 GAT, 2x2x16 JKNet, 64 MLP), and a $0.5$ dropout rate. 
We fix an update step budget of $200$ per task and use Adam~\cite{Adam} to optimize cross-entropy. We implemented GAT, GraphSAGE-mean, Simplified GCN, and JKNet with \textit{dgl}~\cite{dgl} and use \textit{torch-geometric}~\cite{geometric} for GraphSAINT.
We had to disable GraphSAINT's norm (re-)computation for each task such that our experiments could finish in reasonable time.

After hyperparameter optimization, we apply the models on all datasets using different history sizes.
We run our incremental training method for graph learning from Section~\ref{sub:incrementaltraining} for each of the models under both the warm restart and the cold restart configuration.
Finally, we conduct an ablation study to isolate the effect of incremental training.
All experiments are repeated 10 times with different random seeds.

\subsection{Evaluation Measures}
Our primary evaluation measure for our models $f$ is accuracy.
With $\operatorname{acc}_t(f^\embrace{t})$, we denote the accuracy of model $f^\embrace{t}$ on task $\gT_t$. 
We aggregate accuracy scores over the sequence of tasks $\gT_1, \ldots, \gT_T$ by using their unweighted average~\cite{DBLP:conf/nips/Lopez-PazR17}:
\begin{equation*}
    \operatorname{acc}(f) = \frac{1}{T} \sum_{t \in 1, \ldots, T} \operatorname{acc}_t(f^\embrace{t})
\end{equation*}
Following Lopez-Paz \& Ranzato~\cite{DBLP:conf/nips/Lopez-PazR17}, we use Forward Transfer (FWT) to quantify the effect of reusing previous parameters.
This is reflected by the accumulated differences in accuracy between the $f_\text{warm}$ and $f_\text{cold}$ models, defined below:
\begin{equation*}
  \operatorname{FWT}(f_\text{warm}, f_\text{cold}) = \frac{1}{T-1} \sum_{t \in 2, \ldots,  T} \operatorname{acc}_t(f_\text{warm}^\embrace{t}) - \operatorname{acc}_t(f_\text{cold}^\embrace{t})
\end{equation*}

\section{Experimental Results}\label{sec:experiments}
\label{sec:results} 
Table~\ref{tab:results} shows the aggregated results of \numevalsteps{} evaluation steps (48 configurations with 10 repetitions on two datasets with 12 tasks each and one dataset with 18 tasks). 
We consider method $A$ better than method $B$ when the mean accuracy of $A$ is higher than the one of $B$ and the 95\% confidence intervals do not overlap~\cite{deeplearningbook}.
In terms of absolute best methods per setting (= dataset $\times$ history size), we find that GraphSAGE consistently yields the highest scores except for \dblphard, where it is challenged by Simplified GCN. 

Regarding the comparison of history sizes (\ie \textit{explicit knowledge}, see introduction), the highest scores are achieved in almost all cases by using an unlimited history size, \ie using the full graph's history.
However, on all datasets, the scores for training with limited window sizes larger than 1 are close to the ones of full-graph training.
With history sizes that cover 50\% of the GNN's receptive field, all methods achieve at least 95\% relative accuracy compared to the same model under full-history training.
When 75\% of the receptive field is covered, the models yield at least 99\% relative accuracy. To compute these percentages, we have selected the better one of either cold or warm restarts for each method.

Regarding the influence of \emph{implicit knowledge}, we find that reusing parameters (warm restarts) leads to notably higher scores than re-training from scratch, when the history size comprises only one single previous task (see column Forward Transfer).
The average Forward Transfer across all models and datasets with history size $c=1$ is five accuracy points.

Regarding isotropic vs anisotropic GNNs, we find that GAT and GraphSAGE perform similarly well on \dblpeasy (on which the learning rate was tuned). However, GraphSAGE-mean yields higher scores on \dblphard and \wos, which might indicate that GraphSAGE-mean is more robust to hyperparameters than GAT.

Regarding memory-efficient methods, we observe that the scores of Simplified GCN are among the highest of all methods on \dblphard. 
To understand this result, we recall that Simplified GCN uses only one single weight matrix of shape $n_\text{features} \times n_\text{outputs}$, which leads to 300k learnable parameters on \dblphard, but only 27k and 34k on \dblpeasy and \wos, respectively. For comparison, GraphSAGE has 146k learnable parameters on \dblpeasy, 264k on \dblphard, and 310k on \wos.
On the other hand, GraphSAINT yields high scores on \wos, comparable to GraphSAGE, but lower scores on both DBLP datasets.

\begin{table*}
\centering
\caption{Accuracy (with 95\% confidence intervals (CI) via 1.96 standard error of the mean) and Forward Transfer (averaged difference of warm and cold restarts) on our datasets with different history sizes (column \textbf{c}). The best method per case (= 1 dataset + 1 history size) is marked in bold, along with methods where the 95\% CI overlaps.}  
\label{tab:results}
\begin{tabular}{llccccccccc}
\toprule
& {} & \multicolumn{3}{c}{GAT} & \multicolumn{3}{c}{GraphSAGE-Mean} & \multicolumn{3}{c}{MLP (Baseline)} \\
          & {} &     \multicolumn{2}{c}{avg. accuracy} & FWT &     \multicolumn{2}{c}{avg. accuracy} & FWT &   \multicolumn{2}{c}{avg. accuracy} & FWT \\
          & {} &    cold & warm  & {} &     cold & warm & {}  &   cold & warm & {} \\

\textbf{Dataset} & \textbf{c} &               &          &               &          &               &          \\
\midrule
\multirow{4}{*}{\textbf{\dblpeasy{}}}
        & \textbf{1} & $60.8 \pm 0.5$ & $\mathbf{64.9 \pm 0.4}$ & $+4.5$ & $60.4 \pm 0.5$ & $\mathbf{65.1 \pm 0.4}$ & $+5.2$ & $56.1 \pm 0.4$ & $62.2 \pm 0.5$ & $+6.6$\\
        & \textbf{3} & $\mathbf{68.9 \pm 0.3}$ & $\mathbf{69.3 \pm 0.3}$ & $+0.2$ & $\mathbf{68.7 \pm 0.3}$ & $\mathbf{69.3 \pm 0.3}$ & $+0.7$ & $61.0 \pm 0.5$ & $62.9 \pm 0.4$ & $+2.0$\\
        & \textbf{6} & $\mathbf{70.3 \pm 0.4}$ & $70.2 \pm 0.4$ & $-0.1$ & $\mathbf{71.1 \pm 0.4}$ & $\mathbf{70.9 \pm 0.4}$ & $-0.3$ & $62.7 \pm 0.3$ & $62.7 \pm 0.4$ & $-0.2$\\
     & \textbf{full} & $70.2 \pm 0.4$ & $70.2 \pm 0.4$ & $+0.1$ & $\mathbf{71.6 \pm 0.4}$ & $\mathbf{71.4 \pm 0.3}$ & $-0.2$ & $63.4 \pm 0.3$ & $61.9 \pm 0.4$ & $-1.2$\\
          \midrule
\multirow{4}{*}{\textbf{\dblphard{}}} 
        & \textbf{1} & $39.4 \pm 0.2$ & $39.1 \pm 0.2$ & $-0.1$ & $34.5 \pm 0.4$ & $40.0 \pm 0.2$ & $+5.9$ & $31.6 \pm 0.3$ & $38.3 \pm 0.3$ & $+7.4$\\
        & \textbf{3} & $44.0 \pm 0.2$ & $43.7 \pm 0.2$ & $-0.4$ & $44.3 \pm 0.2$ & $\mathbf{45.1 \pm 0.2}$ & $+0.8$ & $33.7 \pm 0.3$ & $38.9 \pm 0.2$ & $+5.6$\\
        & \textbf{6} & $45.1 \pm 0.3$ & $45.3 \pm 0.3$ & $+0.2$ & $\mathbf{46.5 \pm 0.3}$ & $\mathbf{46.7 \pm 0.3}$ & $+0.2$ & $39.2 \pm 0.2$ & $38.3 \pm 0.2$ & $-0.7$\\
     & \textbf{full} & $45.6 \pm 0.3$ & $45.6 \pm 0.3$ & $-0.1$ & $46.8 \pm 0.2$ & $47.1 \pm 0.3$ & $+0.4$ & $38.2 \pm 0.2$ & $36.7 \pm 0.2$ & $-1.1$\\
          \midrule
\multirow{4}{*}{\textbf{\wos{}}} 
        & \textbf{1} & $61.6 \pm 0.9$ & $65.4 \pm 0.9$ & $+3.8$ & $65.4 \pm 0.9$ & $\mathbf{68.6 \pm 1.0}$ & $+3.3$ & $62.7 \pm 0.9$ & $66.3 \pm 0.9$ & $+3.9$\\
        & \textbf{4} & $64.5 \pm 0.8$ & $65.3 \pm 0.9$ & $+0.9$ & $\mathbf{68.0 \pm 0.8}$ & $\mathbf{69.0 \pm 0.8}$ & $+1.1$ & $66.3 \pm 0.7$ & $65.7 \pm 0.8$ & $-0.7$\\
        & \textbf{8} & $65.1 \pm 0.8$ & $65.4 \pm 0.8$ & $+0.3$ & $\mathbf{68.8 \pm 0.7}$ & $\mathbf{69.0 \pm 0.8}$ & $+0.2$ & $64.2 \pm 0.8$ & $65.3 \pm 0.7$ & $+0.9$\\
     & \textbf{full} & $64.3 \pm 0.8$ & $65.4 \pm 0.8$ & $+0.2$ & $\mathbf{69.0 \pm 0.7}$ & $\mathbf{68.4 \pm 0.7}$ & $-0.7$ & $65.4 \pm 0.8$ & $64.4 \pm 0.6$ & $-1.1$\\
          \midrule\\
          & {} & \multicolumn{3}{c}{Simplified GCN} & \multicolumn{3}{c}{GraphSAINT} & \multicolumn{3}{c}{Jumping Knowledge} \\
          & {} &     \multicolumn{2}{c}{avg. accuracy} & FWT &     \multicolumn{2}{c}{avg. accuracy} & FWT &   \multicolumn{2}{c}{avg. accuracy} & FWT \\
          & {} &    cold & warm  & {} &     cold & warm & {}  &   cold & warm & {} \\
\midrule
\multirow{4}{*}{\textbf{\dblpeasy}} 
        & \textbf{1} & $57.1 \pm 0.4$ & $63.7 \pm 0.4$ & $+7.2$ & $62.1 \pm 0.3$ & $63.2 \pm 0.4$ & $+1.2$ & $56.2 \pm 0.5$ & $61.4 \pm 0.5$ & $+5.6$\\
        & \textbf{3} & $66.4 \pm 0.3$ & $67.4 \pm 0.3$ & $+1.2$ & $66.4 \pm 0.4$ & $65.3 \pm 0.5$ & $-0.9$ & $65.2 \pm 0.3$ & $65.9 \pm 0.5$ & $+1.0$\\
        & \textbf{6} & $69.3 \pm 0.4$ & $69.3 \pm 0.4$ & $+0.1$ & $68.1 \pm 0.4$ & $65.5 \pm 0.7$ & $-2.1$ & $68.0 \pm 0.4$ & $66.9 \pm 0.6$ & $-0.7$\\
     & \textbf{full} & $\mathbf{71.0 \pm 0.4}$ & $70.0 \pm 0.4$ & $-1.0$ & $68.4 \pm 0.5$ & $65.7 \pm 0.5$ & $-2.8$ & $68.7 \pm 0.4$ & $66.3 \pm 0.4$ & $-2.5$\\
          \midrule
\multirow{4}{*}{\textbf{\dblphard}} 
        & \textbf{1} & $34.5 \pm 0.3$ & $\mathbf{41.0 \pm 0.3}$ & $+7.0$ & $35.9 \pm 0.3$ & $35.6 \pm 0.4$ & $+0.5$ & $33.0 \pm 0.2$ & $35.3 \pm 0.3$ & $+2.9$\\
        & \textbf{3} & $44.1 \pm 0.2$ & $\mathbf{44.8 \pm 0.3}$ & $+0.8$ & $39.3 \pm 0.3$ & $38.1 \pm 0.5$ & $-0.6$ & $39.1 \pm 0.3$ & $38.8 \pm 0.4$ & $+0.3$\\
        & \textbf{6} & $\mathbf{46.9 \pm 0.3}$ & $46.2 \pm 0.3$ & $-0.4$ & $40.6 \pm 0.3$ & $38.8 \pm 0.6$ & $-1.2$ & $41.0 \pm 0.3$ & $40.1 \pm 0.5$ & $-0.3$\\
     & \textbf{full} & $\mathbf{48.8 \pm 0.4}$ & $47.5 \pm 0.3$ & $-1.2$ & $41.0 \pm 0.4$ & $40.7 \pm 0.4$ & $-0.3$ & $41.6 \pm 0.3$ & $40.8 \pm 0.2$ & $-0.9$\\
          \midrule
\multirow{4}{*}{\textbf{\wos}} 
        & \textbf{1} & $62.3 \pm 0.9$ & $64.5 \pm 0.8$ & $+2.3$ & $65.7 \pm 0.8$ & $\mathbf{68.6 \pm 0.8}$ & $+3.0$ & $64.1 \pm 0.9$ & $\mathbf{68.3 \pm 0.9}$ & $+4.3$\\
        & \textbf{4} & $64.4 \pm 0.8$ & $64.4 \pm 0.8$ & $-0.0$ & $67.3 \pm 0.8$ & $\mathbf{68.4 \pm 0.7}$ & $+1.0$ & $67.1 \pm 0.8$ & $\mathbf{68.2 \pm 0.8}$ & $+1.1$\\
        & \textbf{8} & $65.3 \pm 0.8$ & $64.0 \pm 0.7$ & $-1.4$ & $\mathbf{68.1 \pm 0.8}$ & $\mathbf{68.0 \pm 0.7}$ & $-0.1$ & $\mathbf{67.8 \pm 0.8}$ & $\mathbf{67.7 \pm 0.7}$ & $-0.3$\\
     & \textbf{full} & $62.4 \pm 0.8$ & $61.7 \pm 0.6$ & $-0.8$ & $\mathbf{68.2 \pm 0.8}$ & $66.1 \pm 0.8$ & $-2.2$ & $66.8 \pm 0.8$ & $64.5 \pm 0.7$ & $-2.6$\\
\bottomrule
\end{tabular}
\end{table*}

\begin{figure*}[ht!]
\centering
  \includegraphics[width=0.9\textwidth]{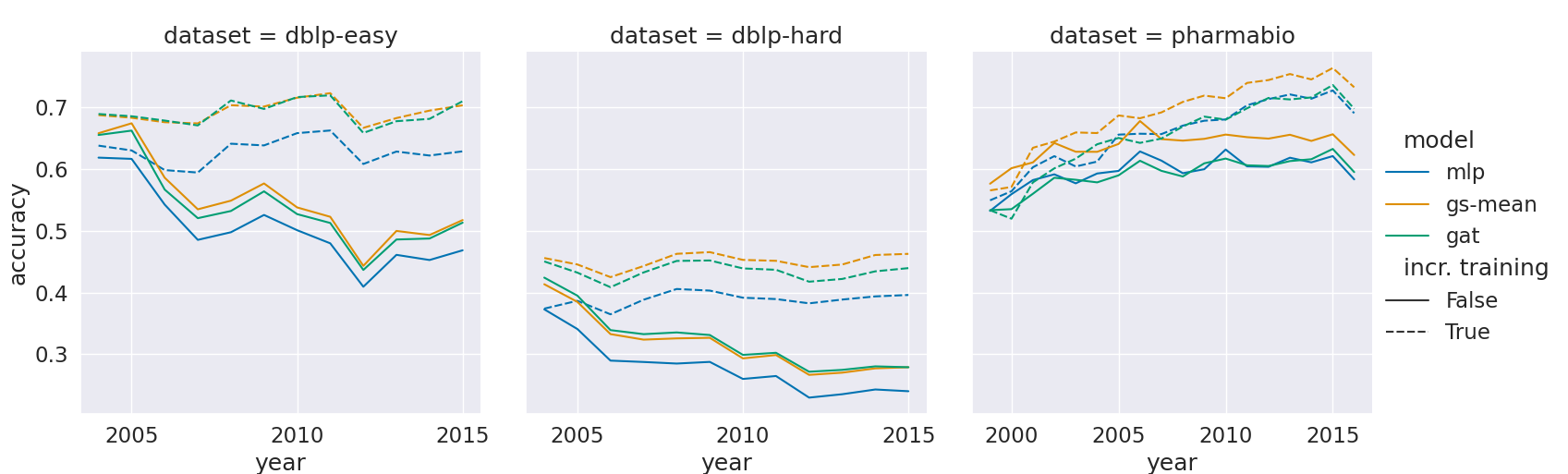}
  \caption{Results of Ablation Study: Accuracy scores of once-trained, static models (solid lines) are lower than incrementally trained models (dashed lines). 
  }\label{fig:exp0}
\end{figure*}

\section{Ablation Study}\label{sec:ablation}
We isolate the effect of incremental training and compare once-trained (static) models against incrementally trained models.
We train the static models for 400 epochs on the data before the first evaluation time step, which comprises 25\% of the total \nodes.
We train incremental models for 200 epochs with history sizes of 3 time steps (4 on the \wos dataset) before evaluating each task. We repeat each experiment 10 times with different random seeds.

In \Figref{fig:exp0}, we see that the accuracy of the static
models decreases over time on \dblpeasy and \dblphard, where new classes appear
over time. On \wos (fixed class set), the accuracy of the static models plateaus,
while the accuracy of incrementally trained models increases.

\section{Discussion}\label{sec:discussion}

We have created a new experimental procedure for lifelong open-world node classification, for which we contribute three newly compiled datasets. In this setup, we have evaluated five representative GNN architectures as well as an MLP baseline under an incremental training procedure. With the goal of generalizable results, we have introduced a new measure for  
$k$-neighborhood time differences $\Dt{k}$, based on which we have selected the history sizes. 

Our main result is that incremental training with limited history sizes is almost as good as using the full history of the graph. 
With window sizes of 3 or 4 (50\% receptive field coverage), GNNs achieve at least 95\% accuracy compared to using all past data for incremental training. With window sizes of 6 or 8 (75\% receptive field coverage), at least 99\% accuracy can be retained. This result holds for standard GNN architectures and also for scalable and sampling-based approaches.
This has direct consequences for 
lifelong learning of GNNs on evolving graphs and, thus, impacts how GNNs can be employed for numerous real-world applications.

We have further investigated on reusing parameters from previous iterations (warm restarts).
Using warm restarts is a viable strategy, \ie reusing an ``old'' model, even though new classes appear during the sequence of tasks.
We have identified a trend that reusing parameters from previous tasks becomes more important when the history sizes are small because less explicit knowledge is available.
Even though it was not our main objective to compare the absolute performances of the models, it is noteworthy that Simplified GCNs perform surprisingly well on DBLP-hard. There, the model yields the highest absolute scores, on par with GraphSAGE-mean,
despite the simplicity of the approach.

A limitation of the present work is that we assume that the labels, \ie the ground truth of each task becomes available as training data for the next task. 
However, in practice only a small fraction of \nodes might come with true labels, \eg if the labels are professionally annotated subjects. 
Investigating whether only a small fraction of annotated nodes are sufficient for retraining would be an interesting direction of future work. 
Furthermore, our method does not yet include an unsupervised mechanism for self-detection of new classes~\cite{DBLP:journals/tkde/MasudGKHT11,DBLP:conf/kdd/FeiW016,DBLP:conf/icdm/WuPZ20}. 
However, our experimental apparatus lays the foundation for adding methods of self-detection of new classes.

To set our work in the broader context of lifelong learning, we reconsider the
gradient episodic memory framework~\cite{DBLP:conf/nips/Lopez-PazR17} for image data, in which examples are independent. 
To cast graph data into independent examples for \node classification, certain preprocessing steps are required such as transforming each \node into a graph~\cite{DBLP:journals/corr/abs-2009-00647}.
This increases the number of inference steps by $\mathcal{O}(|V|)$ compared to our approach. 
Furthermore, we have shown that our approach of incremental training can be applied to various GNN architectures and is orthogonal to sampling and preprocessing approaches. 
In general, our incremental training procedure can be applied to any GNN architecture with few caveats: 
If the GNN architecture relies on transductive learning, the constraints also need to be satisfied during incremental training. 
Similarly, any pre-computation steps such as computing normalizing constants have to be performed again when adapting the model to a new task.


\section{Conclusion}

We have introduced an incremental training scheme for lifelong learning on evolving graphs where new classes appear over time. 
We have proposed a method to select the history sizes for training the models agnostic to different granularities of temporal intervals.
Our analysis of implicit and explicit knowledge in lifelong learning on graphs shows that high levels of accuracy can be preserved, even when training only on a fraction of past data.
Furthermore, using warm restarts becomes more important when few past data is available. 

\textbf{\dataavailabilityandreproducibility}\\
We published our lifelong graph learning datasets (\href{https://zenodo.org/record/3764770#.YCQUOHWYXmg}{zenodo.org/record/3764770\#.YCQUOHWYXmg}) along with an implementation of our experimental procedure (\href{https://github.com/lgalke/lifelong-learning}{github.com/lgalke/lifelong-learning}).

\textbf{\acknowledgments}:
Parts of this research were carried out on the computing infrastructure \emph{bwCloud} of the State of Baden-Würtemberg and the Bioinformatics and Systems Biology group at Ulm University.

\bibliography{main}
\bibliographystyle{IEEEtran}

\end{document}

%% file: math_commands.tex
\usepackage{amsmath,amsfonts,bm}

\def\Figref#1{Figure~\ref{#1}}

\def\Secref#1{Section~\ref{#1}}

\def\eqref#1{equation~\ref{#1}}

\def\1{\bm{1}}

\def\vh{{\bm{h}}}

\def\vy{{\bm{y}}}

\def\mS{{\bm{S}}}

\def\mU{{\bm{U}}}

\def\mX{{\bm{X}}}

\DeclareMathAlphabet{\mathsfit}{\encodingdefault}{\sfdefault}{m}{sl}
\SetMathAlphabet{\mathsfit}{bold}{\encodingdefault}{\sfdefault}{bx}{n}

\def\gG{{\mathcal{G}}}

\def\gK{{\mathcal{K}}}

\def\gT{{\mathcal{T}}}

\def\gV{{\mathcal{V}}}

\def\sY{{\mathbb{Y}}}

